\documentclass[10pt,conference]{IEEEtran}
\usepackage{graphicx}
\usepackage{booktabs}
\usepackage{multirow}
\usepackage[dvipsnames,table,xcdraw]{xcolor}
\usepackage{cite}
\usepackage{textcomp}
\usepackage{bm}
\usepackage{url}
\usepackage{float}
\usepackage{placeins}
\usepackage{ragged2e}
\usepackage{etoolbox}
\usepackage{amssymb}
\usepackage[cmex10]{amsmath}
\usepackage{amsthm}
\usepackage{array}
\usepackage[caption=false,font=footnotesize]{subfig}
\usepackage[hidelinks]{hyperref}
\definecolor{cinza}{RGB}{230,230,230}
\definecolor{DarkBlue}{RGB}{0,0,139}
\definecolor{DarkGreen}{RGB}{0,100,0}
\definecolor{DarkRed}{RGB}{139,0,0}
\newcommand{\eg}{e.g., }

\def\baselinestretch{.85}
\makeatletter
\def\ps@IEEEtitlepagestyle{
  \def\@oddhead{\hfil\normalfont\small\itshape Preprint of a paper accepted at SIBGRAPI 2026.\hfil}
  \let\@evenhead\@oddhead
  \def\@oddfoot{}
  \let\@evenfoot\@oddfoot
}
\makeatother

\begin{document}
\title{Region-Aware CLS Token Augmentation \\ for Fine-Grained Image Retrieval}

\author{\IEEEauthorblockN{Ian de Holanda Cavalcanti Bezerra\IEEEauthorrefmark{1},
Vivek Trivedy\IEEEauthorrefmark{2},
Lucas Pascotti Valem\IEEEauthorrefmark{1}, and
Longin Jan Latecki\IEEEauthorrefmark{2}}
\IEEEauthorblockA{\IEEEauthorrefmark{1}Institute of Mathematics and Computer Science (ICMC),
University of São Paulo (USP), São Carlos, Brazil}
\IEEEauthorblockA{\IEEEauthorrefmark{2}Department of Computer and Information Science,
Temple University, Philadelphia, PA, USA}
\IEEEauthorblockA{Email: idhcb.ian@usp.br, vivek.trivedy@temple.edu,
lucas@icmc.usp.br, latecki@temple.edu}}

\maketitle

\begin{abstract}
Image retrieval methods often rely on a single global semantic descriptor extracted from an image, e.g., the \texttt{[CLS]} token in vision transformers. However, trying to squeeze all the semantic information of an image into a single descriptor can hurt downstream retrieval performance, especially for fine-grained retrieval tasks. In this work, we augment the semantic tokens in the newer visual transformers, the global \texttt{[CLS]} token and the four register tokens, with a carefully selected collection of spatial tokens, aiming to capture the spatial region representation that characterizes the contents captured in each of the semantic tokens. We leverage the DINOv2-reg model, which includes register tokens that emergently learn object and part-based representations. For each ``cue" token (\texttt{[CLS]} and each register token), we find a ``buddy" image patch token and extract an $N \times N$ patch region to produce a set of localized ROI tokens.  Our approach automatically captures important regions of interest without any external bounding boxes or saliency modules, purely by matching semantic tokens with their spatial representation regions. Furthermore, we incorporate these tokens into a multi-vector retrieval framework inspired by ColBERT, enabling fine-grained matching via a per-token alignment mechanism while avoiding the large storage cost of keeping all patch embeddings. Through extensive experiments,
we find that (1) register tokens encode useful fine-grained details that can complement the \texttt{[CLS]} token; (2) automatically pooled ROI tokens further improve fine-grained discrimination; and (3) multi-vector retrieval with a small set of tokens improves over a DINOv2-reg single-vector baseline while remaining tractable for large-scale search. The code is available at \url{https://github.com/IdhcbIan/Augmenting_CLS_with_ROI_tokens}.
\end{abstract}

\section{Introduction} \label{sec:intro}
The task of image retrieval aims to retrieve the $k$ most relevant images from a database for a given query, often by comparing learned semantic global descriptors in a shared embedding space~\cite{multitoken_survey}. With large-scale vision transformers (ViTs)~\cite{dosovitskiy2020image,dinov2}, a common practice is to prepend a token at position zero of the patch sequence so that, after the self-attention layers, it attends to all parts of the image and represents the global semantic content with a single token (the \texttt{[CLS]} token). While effective for broad semantic matching, a single vector may struggle to capture the subtle, localized details crucial for \emph{fine-grained} retrieval~\cite{WANG2024104030}. Such limitations follow from fundamental theoretical constraints on the representational capacity of single-vector embeddings, which cannot express all possible retrieval relations regardless of model size or training data~\cite{weller2025theoretical}.

An intuitive strategy for capturing richer details is to use multiple tokens per image~\cite{multitoken_survey,colbert}. In NLP, ColBERT~\cite{colbert} matches all query tokens against those of the database, improving fine-grained retrieval. However, comparing entire sets of image patch tokens is computationally and storage-wise prohibitive.

Classical region-of-interest (ROI) methods relied on keypoint detectors such as SIFT~\cite{lowe2004distinctive} or saliency maps, and later on CNN-based part detection and attention~\cite{Noh2017delf,Cao2020delg} that highlight discriminative regions without bounding-box annotations. Vision transformers add a new perspective: self-attention and special tokens (\eg \texttt{[CLS]}) reveal local semantics, segmenting objects or parts even without explicit supervision~\cite{caron2021emerging,dinov2}.

\begin{figure}[!ht]
    \centering
    \includegraphics[width=0.46\textwidth]{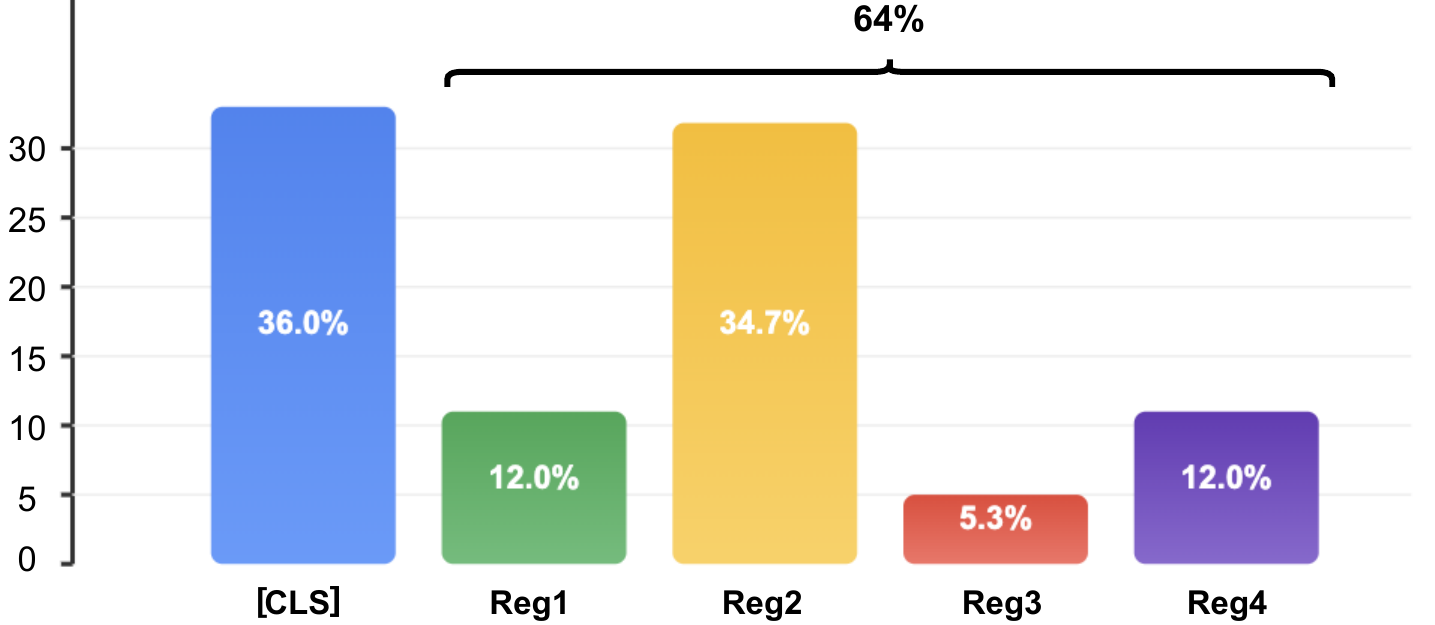}
        \caption{Distribution of most similar tokens of the database images for any given \texttt{[CLS]} token from queries of images in COCO's dog class~\cite{coco}. Interestingly, a \texttt{[CLS]} token from a query image is often most similar to one of the four register tokens of other images (64\%), not their \texttt{[CLS]} token (36\%).}
    \label{fig:token_flow}
\end{figure}

In line with these developments, we propose an efficient multi-vector retrieval framework that automatically discovers discriminative local regions within a single forward pass. We build on DINOv2-reg~\cite{dinov2-reg}, which augments the ViT with \emph{register tokens}: extra learnable tokens that act as ``attention sinks'', absorbing global context so that patch embeddings remain faithful to their local regions. The original authors discard these tokens after training; we instead retain them as additional descriptors that capture object- and part-level features. Fig.~\ref{fig:token_flow} motivates this choice: even among images of the same class, a query \texttt{[CLS]} token is frequently more similar (by dot product) to a \emph{register} token of another image than to that image's \texttt{[CLS]} token.

That register tokens carry part-level semantics exploitable for retrieval is a non-obvious finding that challenges prevailing practice, in which they are typically overlooked in downstream use~\cite{dinov2-reg}. This is analogous to how ColBERT~\cite{colbert} repurposed BERT's existing token representations for late interaction rather than designing new encoders. Crucially, unlike approaches that add external modules~\cite{Noh2017delf,Cao2020delg}, learned queries~\cite{detrex}, or explicit supervision, register tokens are already produced by the standard forward pass at a small inference cost. Concurrent work corroborates this: register neurons encode meaningful information even without register training~\cite{jiang2025visiontransformersdontneed}, and DINOv3~\cite{simeoni2025dinov3} keeps registers as a core architectural element. 

\begin{figure*}[t]
    \centering
        \includegraphics[width=.84\textwidth, trim={0cm 0cm 0cm 5.1cm}, clip]{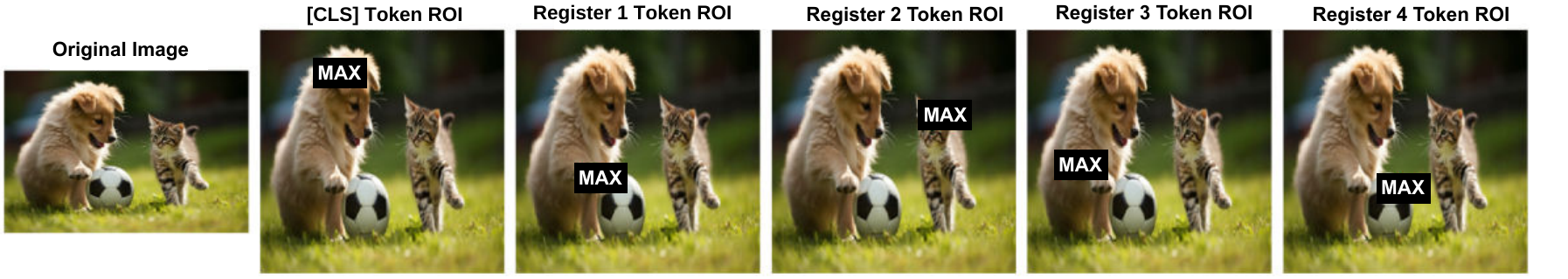}
    \caption{For each cue token (the \texttt{[CLS]} and register tokens), the location of its corresponding buddy token is shown by the black max box. They focus on different parts of the semantic input: dog's head, dog's paw, cat's ear, and ball.}
            \label{fig:attn_viz}
\end{figure*}

We further extend each image's token set with a small number of \emph{ROI tokens}, each summarizing a localized region around a ``cue'' token (the \texttt{[CLS]} and register tokens). For every cue token we locate its most similar patch, its ``buddy'', and mean-pool a small region around it, capturing parts such as a bird's head or a car's headlight without any external proposals, bounding boxes, or saliency modules (Section~\ref{subsec:buddy_tokens}).
Fig.~\ref{fig:attn_viz} illustrates the selected buddy tokens, which focus on
different parts of the semantic input.
Our experiments show that matching registers alongside the \texttt{[CLS]} token
already improves retrieval, and that adding ROI tokens improves it further
(Section~\ref{sec:results}).

For matching, we adopt a per-token, max-sim alignment inspired by ColBERT~\cite{colbert}; because each image carries only $\sim$10 tokens, the similarity computation and storage remain tractable, and retrieval can be further accelerated with single-vector approximations of multi-vector search~\cite{dhulipala2024muveramultivectorretrievalfixed} (Section~\ref{subsec:colbert_training}).

Our \textbf{main contributions} are summarized as follows:
\begin{itemize}
\item A buddy token mechanism conditioned on the cue tokens that exploits the existing DINOv2-reg representations to select and spatially ground a compact set of ROI descriptors. These descriptors are then combined using matching inspired by ColBERT. The resulting representation improves over the corresponding baseline that uses only a single vector while limiting computational and memory costs.
\item Observation and experiments showing that a small number of cue-conditioned ROI tokens can improve retrieval accuracy, particularly for fine-grained class differences. These ROI tokens are selected from internal DINOv2-reg representations without external proposals or manual bounding boxes.
\item Observation and experiments showing that register tokens carry meaningful information for image retrieval.
\end{itemize}

\section{Related Work}

\label{sec:related}
Early retrieval relied on local keypoints (e.g., SIFT~\cite{lowe2004distinctive}) or saliency, in hand-crafted pipelines separate from the embedding model. DELF (Deep Local Features)~\cite{Noh2017delf} instead produces CNN-based local descriptors with an attention mechanism selecting useful features, and DELG (Deep Local and Global Features)~\cite{Cao2020delg} adds a dual head yielding both a global embedding and attentive local descriptors. Vision transformers (ViTs) broadened ROI discovery: the \texttt{[CLS]} token focuses on objects or parts~\cite{caron2021emerging}, while register tokens or extra queries capture distinct regions~\cite{dinov2-reg,detrex}, and DEtection TRansformer (DETR)-like models learn bounding boxes via attention without proposal networks. Some methods instead enhance retrieval through learned similarity: the Reranking Transformer (RRT)~\cite{Tan2021} predicts pairwise similarity from global and local descriptors, replacing costly geometric verification, while AMES (Asymmetric and Memory-Efficient Similarity Estimation)~\cite{Suma_2024_ECCV} uses \textit{asymmetric} estimation, representing database images with far fewer compact binary descriptors than queries.

For global representations, the common ViT descriptor is the \texttt{[CLS]} token~\cite{dosovitskiy2020image}, though mean-pooling patch tokens performs comparably~\cite{Song2023dtop}, showing summary information is distributed across tokens. Other works combine multiple layers or concatenate the \texttt{[CLS]} token with pooled patches~\cite{globallocal,Song2023dtop}, or \emph{weight and select} patches by importance to discard uninformative backgrounds~\cite{globallocal}, a partial step toward multi-vector retrieval that typically boosts fine-grained discrimination.

Multi-vector retrieval, however, is costly: token clustering or centering compresses ColBERT-style indexes with little accuracy loss~\cite{Lupart2024token}, and quantization or dimensionality reduction manage index size~\cite{Santhanam2023ann}, while others reduce the token count within the model, as DynamicViT~\cite{Rao2021dynamicvit} prunes low-importance patches and TokenLearner~\cite{Ryoo2021tokenlearner} merges patches into fewer learned tokens. Our approach follows this philosophy, using the model's own similarity scores to keep the token set compact and discriminative.

Finally, fine-grained image retrieval (FGIR) targets the \emph{same subordinate category} (e.g., bird species or car model), challenging due to subtle inter-class and large intra-class variation~\cite{sop}. Part-based strategies that model and match key regions improve retrieval of similar objects~\cite{Noh2017delf,inshop}, while attribute or hierarchical labels guide focus toward class-specific features~\cite{proxyncaplus}. Our method follows these perspectives via register tokens capturing object parts, enriched by ROI tokens for patch-level detail.

\section{Methodology}
\label{sec:methodology}

\begin{figure}[h!]
    \centering
    \includegraphics[width=0.49\textwidth]{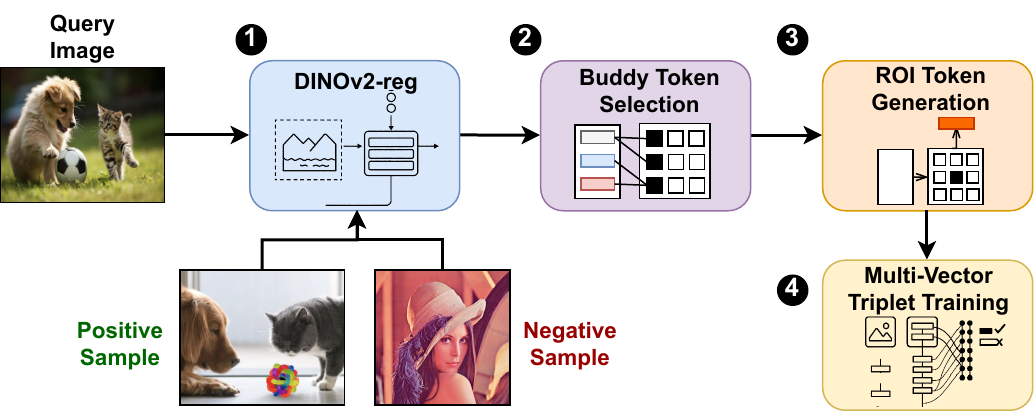}
    \caption{The data flow of our training setup. First, the \texttt{[CLS]} and register tokens are extracted with DINOv2-reg. Second, for each of these cue tokens, the buddy token is identified as the most similar image patch token. Third, the ROI token is constructed by mean pooling the buddy token and the image patch tokens in the region around it. In our experiments, we use only 10 tokens per image. Finally, the standard triplet training is performed with ColBERT-style similarity between image pairs.}
    \label{fig:data_flow}
\end{figure}

The overall workflow of our approach is illustrated in Fig.~\ref{fig:data_flow}.
We first describe our DINOv2-reg backbone and the tokens it produces (Section~\ref{subsec:model_forward}), then present our new ROI token approach (Section~\ref{subsec:buddy_tokens}), followed by the ColBERT-style multi-token loss (Section~\ref{subsec:colbert_training}).

\subsection{Model and Forward Pass} \label{subsec:model_forward} Our backbone is the base DINOv2-reg~\cite{dinov2-reg}, a vision transformer that takes an image $I \in \mathbb{R}^{H \times W \times 3}$ as input. After patch embedding and transformer layers, it outputs a sequence of tokens: \begin{equation}
\mathbf{T} = \big[\mathbf{t}_{\text{cls}}, \mathbf{t}_{r}^{1}, \dots, \mathbf{t}_{r}^{R}, \mathbf{t}_p^1, \dots, \mathbf{t}_p^P\big] \in \mathbb{R}^{(1+R+P)\times d},
\end{equation}
where $\mathbf{t}_{\text{cls}}$ is the global \texttt{[CLS] token}; $\mathbf{t}_{r}^{i}$ for $i = 1, \ldots, R$ are the \emph{register tokens}; and $\mathbf{t}_p^j$ for $j = 1, \ldots, P$ are the image patch tokens.
We aim to build a small set of final tokens for retrieval, balancing global context and fine-grained local detail.

 \subsection{Regions-of-Interest (ROI) Tokens} \label{subsec:buddy_tokens}
 Our motivation is to use internal model representations to automatically identify and use useful regions within an image.  While \texttt{[CLS]} and register tokens capture broad or part-level features, we want each cue token to ``zoom in" on a specific \emph{region of interest} that it finds relevant. Naively incorporating all P patch tokens into retrieval would significantly increase the storage and computation required for indexing and search.  Instead of storing \emph{all} patch tokens, we extract a small number of image patch tokens, which are then grouped into ROIs.

 \paragraph{Buddy ROI Token}
 For each cue token $\mathbf{t}_{\text{cue}} \in \{\mathbf{t}_{\text{cls}}, \mathbf{t}_r^1, \ldots, \mathbf{t}_r^R\}$, we perform the following steps:
 \begin{enumerate}
 \item \textbf{Locate the buddy patch}: Compute dot-product similarities: \[ s_j = \mathbf{t}_{\text{cue}}^\top \mathbf{t}_p^j,\quad j=1,\ldots,P, \] and find the patch index $j^*$ with highest similarity $s_{j^*}$.

 \item \textbf{Mean-pool to ROI token}:
 Let $\Omega$ be an $N\times N$ region of patches around patch $j^*$, where $N=3$ is our default setting.
 We average those patch embeddings into a single-vector: \begin{equation}
 \mathbf{b}_{\text{cue}} = \frac{1}{|\Omega|}\sum_{\mathbf{t}_p \in \Omega}\mathbf{t}_p. \label{eq:buddy_avg}
 \end{equation}
This yields a token capturing a small local region near the highest-similarity patch. We choose mean-pooling over max-pooling or attention-weighted aggregation because it adds no learnable parameters and acts as a spatial smoothing operator that denoises individual patch embeddings.
Near the grid boundary, $\Omega$ is clamped to the valid extent. For INSTRE, padded positions introduced by the multi-scale padding augmentation are also masked out, as detailed in Section~\ref{subsec:implementation_details}.

\end{enumerate}

\noindent We gather these ROI tokens across the \texttt{[CLS]} and register tokens, automatically highlighting localized parts the cue token deems most relevant.

 \noindent \paragraph{Resulting Token Set} For $R$ register tokens plus the \texttt{[CLS]} token, the final retrieval set is: \begin{equation} \mathbf{E}(I) = \underbrace{\{\mathbf{t}_{\text{cls}}, \mathbf{t}_r^1, \dots, \mathbf{t}_r^R\}}_{\text{cue tokens}} \;\cup\; \{\mathbf{b}_{\text{cls}}, \mathbf{b}_r^1, \dots, \mathbf{b}_r^R\}, \end{equation} i.e., $1 + R$ cue tokens and $1 + R$ ROI tokens, for a total of $2(R+1)$. If $R=4$, that is 10 tokens total.

 \subsection{Multi-Vector Training} \label{subsec:colbert_training} We train this multi-vector representation using a \emph{max-sim} formulation inspired by ColBERT~\cite{colbert}. Let $\mathbf{E}(Q)=\{\mathbf{q}_1,\ldots,\mathbf{q}_m\}$ be $m$ tokens for a query image $Q$, and $\mathbf{E}(D)=\{\mathbf{d}_1,\ldots,\mathbf{d}_n\}$ be $n$ tokens for database image $D$. We define:
 \begin{equation} S(Q, D) = \sum_{i=1}^m \max_{1 \leq j \leq n}\,(\mathbf{q}_i^\top \mathbf{d}_j).
 \end{equation}
  We also L2 normalize each token.

 During training, we use a triplet-like objective: for each triplet $(Q, D^+, D^-)$,
 \begin{equation}
 \mathcal{L} = \sum_{\text{triplet}} \Big[ \max\big(0,\; \alpha + S(Q, D^-) - S(Q, D^+)\big)\Big],
 \end{equation}
where $D^+$ is an image from the same class as query $Q$, $D^-$ is from a different class, and $\alpha$ is the margin. This pushes the score $S(Q, D^+)$ above $S(Q, D^-)$ by at least $\alpha$, ensuring that each query token better aligns with tokens of the positive image.

\paragraph{Inference} At inference, buddy-token extraction reuses a single backbone pass and adds only two operations: \emph{(i)} one similarity map between each of the $R+1$ cue tokens and the patch grid, followed by $\arg\max$; and \emph{(ii)} an $N\times N$ mean pooling around the selected patch. Since these operations introduce no additional parameters or forward passes, the encoding cost remains dominated by the backbone. Notice that the representation has $2(R+1)=10$ tokens regardless of the patch-grid size, so that \emph{max-sim} evaluates $10\times10=100$ token similarities per image pair, compared with $1$ for a single vector and $201^2=40{,}401$ for the full-token representation. Candidate images are then ranked according to $S(Q,D)$. For further scalability, the \texttt{[CLS]} token can retrieve an approximate nearest-neighbor (ANN) shortlist, which is subsequently re-ranked using $S(Q,D)$. For a fixed shortlist size, the number of candidates evaluated by \emph{max-sim} does not grow with the database size. The integration and benchmarking of this two-stage solution remain future work.

\begin{table*}[t]
\caption{Recall@k metrics on CUB-200, In-Shop, Cars-196, and Stanford Online Products. The table compares reported methods with different backbones, embedding dimensions, and training objectives; our DINOv2-reg ablations isolate the contribution of the proposed token selection and multi-vector scoring.}
\centering
\resizebox{\textwidth}{!}{\begin{tabular}{l|l|l|llll|llll|llllllll}
\toprule
\textbf{Method}       & \textbf{Dim.} & \textbf{Architecture} & \multicolumn{4}{c|}{\textbf{CUB-200}} & \multicolumn{4}{c|}{\textbf{In-Shop}} & \multicolumn{4}{c}{\textbf{Cars-196}}                   & \multicolumn{4}{c}{\textbf{Stanford Online Products}} \\
             &     &              & 1     & 2     & 4     & 8    & 1     & 10    & 20    & 30   & 1    & 2    & 4    & \multicolumn{1}{l|}{8}    & 1         & 2        & 4        & 8        \\ \midrule
NSoftmax ~\cite{nsoftmax}     & 512 & ResNet50          & 61.3  & 73.9  & 83.5  & 90   & 86.6  & 97.5  & 98.4  & 98.8 & 84.2 & 90.4 & 94.4 & \multicolumn{1}{l|}{96.9} & 78.2      & 90.6     & 96.2     & -        \\
ProxyNCA++ ~\cite{proxyncaplus}  & 512 & ResNet50          & 69.0  & 79.8  & 87.3  & 92.7 & 90.4  & 98.1  & 98.8  & 99.0 & 86.5 & 92.5 & 95.7 & \multicolumn{1}{l|}{97.7} & 80.7      & 92.0     & 96.7     & 98.9     \\
A-BIER ~\cite{abier}      & 512 & GoogleNet    & 57.5  & 68.7  & 78.3  & 86.2 & 93.1  & 95.1  & 96.9  & 97.5 & 82.0 & 89.0 & 93.2 & \multicolumn{1}{l|}{96.1} & 74.2      & 86.9     & 94.0     & 97.8     \\
ABE ~\cite{abe}         & 512 & GoogleNet    & 60.6  & 71.5  & 79.8  & 87.4 & 87.3  & 96.7  & 97.9  & 98.2 & 85.2 & 90.5 & 94.0 & \multicolumn{1}{l|}{96.1} & 76.3      & 88.4     & 94.8     & 98.2     \\
SM  ~\cite{sm}         & 512 & GoogleNet    & 56.0  & 68.3  & 78.2  & 86.3 & 90.7  & 97.8  & 98.5  & 98.8 & 83.4 & 89.9 & 93.9 & \multicolumn{1}{l|}{96.5} & 75.3      & 87.5     & 93.7     & 97.4     \\
Proxy-Anchor ~\cite{proxy_anchor} & 512 & Inception-BN & 68.4  & 79.2  & 86.8  & 91.6 & 91.5  & 98.1  & 98.8  & 99.1 & 86.1 & 91.7 & 95.0 & \multicolumn{1}{l|}{97.3} & 79.1      & 90.8     & 96.2     & 98.7     \\
SoftTriple ~\cite{softtriple}   & 512 & Inception-BN & 65.4  & 76.4  & 84.5  & 90.4 & -     & -     & -     & -    & 84.5 & 90.7 & 94.5 & \multicolumn{1}{l|}{96.9} & 78.6      & 86.6     & 91.8     & 95.4     \\
HORDE  ~\cite{horde}      & 512 & Inception-BN & 66.8  & 77.4  & 85.1  & 91.0 & 90.4  & 97.8  & 98.4  & 98.7 & 86.2 & 91.9 & 95.1 & \multicolumn{1}{l|}{97.2} & 80.1      & 91.3     & 96.2     & 98.7     \\
XBM  ~\cite{xbm}        & 512 & Inception-BN & 65.8  & 75.9  & 84.0  & 89.9 & 89.9  & 97.6  & 98.4  & 98.6 & 82.0 & 88.7 & 93.1 & \multicolumn{1}{l|}{96.1} & 79.5      & 90.8     & 96.1     & 98.7     \\
MS ~\cite{ms}          & 512 & Inception-BN & 65.7  & 77.0  & 86.3  & 91.2 & 89.7  & 97.9  & 98.5  & 98.8 & 84.1 & 90.4 & 94.0 & \multicolumn{1}{l|}{96.5} & 78.2      & 90.5     & 96.0     & 98.7     \\
HTL  ~\cite{htl}        & 512 & Inception-BN & 57.1  & 68.8  & 78.7  & 86.5 & 80.9  & 94.3  & 95.8  & 97.2 & 81.4 & 88.0 & 92.7 & \multicolumn{1}{l|}{95.7} & 74.8      & 88.3     & 94.8     & 98.4     \\ \midrule
IRT\_R ~\cite{irtr}      & 384 & DeiT-S       & 76.6  & 85.0  & 91.1  & 94.3 & 91.9  & 98.1  & 98.7  & 98.9 & -    & -    & -    & \multicolumn{1}{l|}{-}    & 84.2      & 93.7     & 97.3     & 99.1     \\
Sph-DeiT ~\cite{hyperbolic}   & 384 & DeiT-S       & 76.2  & 84.5  & 90.2  & 94.3 & 89.6  & 97.2  & 98.0  & 98.4 & 81.7 & 88.6 & 93.4 & \multicolumn{1}{l|}{96.2} & 82.5      & 92.9     & 97.1     & 99.1     \\
Sph-DINO ~\cite{hyperbolic}    & 384 & ViT          & 78.7  & 86.7  & 91.4  & 94.9 & 90.1  & 97.1  & 98.0  & 98.4 & 86.6 & 91.8 & 95.2 & \multicolumn{1}{l|}{97.4} & 82.2      & 92.1     & 96.8     & 98.9     \\
Sph-ViT  ~\cite{hyperbolic}    & 384 & ViT(IN21k)   & 85.1  & 90.7  & 94.3  & 96.4 & 90.4  & 97.4  & 98.2  & 98.6 & 81.7 & 89.0 & 93.0 & \multicolumn{1}{l|}{95.8} & 82.1      & 92.5     & 97.1     & 99.1     \\
Hyp-DeiT ~\cite{hyperbolic}    & 384 & DeiT-S       & 77.8  & 86.6  & 91.9  & 95.1 & 90.5  & 97.8  & 98.5  & 98.9 & 86.4 & 92.2 & 95.5 & \multicolumn{1}{l|}{97.5} & 83.3      & 93.5     & 97.4     & 99.1     \\
Hyp-DINO ~\cite{hyperbolic}    & 384 & ViT          & 80.9  & 87.6  & 92.4  & 95.6 & 92.4  & \textbf{98.4}  & \textbf{98.9}  & 99.1 & 89.2 & 94.1 & 96.7 & \multicolumn{1}{l|}{98.1} & 85.1      & 94.4     & 97.8     & 99.3     \\
Hyp-ViT ~\cite{hyperbolic}     & 384 & ViT(IN21k)   & 85.6  & 91.4  & 94.8  & 96.7 & 92.5  & 98.3  & 98.8  & 99.1 & 86.5 & 92.1 & 95.3 & \multicolumn{1}{l|}{97.3} & 85.9      & 94.9     & \textbf{98.1}     & \textbf{99.5}     \\ \midrule
\textbf{DINOv2-reg (ours)}   & 384 & ViT  & \textbf{87.4}  & \textbf{92.7}  & \textbf{95.2}  & \textbf{97.2} & \textbf{92.9}  & \textbf{98.4}  & \textbf{98.9}  & \textbf{99.4} & \textbf{90.5} & \textbf{94.9} & \textbf{97.1} & \multicolumn{1}{l|}{\textbf{98.4}} & \textbf{86.2}      & \textbf{95.4}     & 98.0     & \textbf{99.5} \\ \bottomrule
\end{tabular}
}
\label{tab:resultscomparison}
\end{table*}

\section{Experiments}
\label{sec:results}
We evaluate our approach on several popular image retrieval benchmarks, highlighting the improvements offered by register tokens, ROI token selection, and the ColBERT-style multi-vector training. We also provide several ablation testing configurations of our approach.

\subsection{Datasets}
\label{subsec:datasets}
We considered a total of 5 datasets, which are described as follows:
\begin{enumerate}
    \item \textbf{CUB-200} ~\cite{WahCUB_200_2011} is a fine-grained dataset containing 11,788 images covering 200 classes of birds.
    \item \textbf{In-Shop Clothes Retrieval} ~\cite{inshop} is a clothing dataset containing 52,712 total images and 7,896 classes.
    \item \textbf{Cars-196} ~\cite{cars196} is a dataset of car models containing 16,185 total images and 196 classes.
    \item \textbf{Stanford Online Products (SOP)} ~\cite{sop} contains 120,053 images of 22,634 products downloaded from \textit{eBay.com}.
    \item \textbf{INSTRE} \cite{paperINSTRE} is a challenging benchmark for instance-level image retrieval. It is widely used to test retrieval when targets are small in the image and scenes are cluttered. It contains 28,543 images depicting objects such as toys, books, and logos with natural backgrounds. The dataset is partitioned into 27,293 images for retrieval (gallery set) and 1,250 images as queries~\cite{paperINSTREProtocol}.
\end{enumerate}

In training, we used the default training and testing splits provided by each dataset.
For INSTRE, the gallery set was used for training.
Our splits follow the common approaches used in ~\cite{irtr,hyperbolic}.
Following standard practice, we report Recall@K. For CUB, SOP, and Cars-196, we use $K=\{1,2,4,8\}$. For In-Shop, we use $K=\{1,10,20,30\}$.

\subsection{Implementation Details}
\label{subsec:implementation_details}

We start off with the same model across all experiments, the pretrained DINOv2-reg-base from Meta~\cite{dinov2-reg}, which includes 4 register tokens. We train the model on the training split using our multi-vector Triplet ColBERT Loss over the generated semantic and ROI tokens, backpropagating through the hard buddy pooling argmax selection of the ROI with a steep softmax in the backward pass. This allows us to train the whole vision encoder and its ROI selection process on each dataset. 

We employ standard preprocessing techniques, using an image size of 224; additional data augmentation was used in cases where the dataset applies cropping to its queries.
For evaluation, we compute Recall@K as our primary metric and Mean Average Precision (mAP) in some cases to facilitate comparisons with the literature, especially the INSTRE dataset.

For optimization, we employ AdamW with a learning rate of $10^{-5}$ and weight decay of 0.01. We set the triplet-loss margin to $\alpha=0.5$ for INSTRE and $\alpha=0.3$ for all other datasets. We used a batch size of 512 for the CUB, INSTRE, and Cars datasets, and 1024 for the larger ones. The number of training steps was adjusted based on dataset complexity: 1,000 steps for CUB and Cars, 1,500 for In-Shop, and 10,000 for SOP.

\paragraph{Triplet Selection}
To construct informative training triplets using hard negative mining, each anchor image in a batch is paired with its least similar match from the same class (the hardest positive) and its most similar non-match from a different class (the hardest negative) based on their similarities in the embedding space.
Focusing on near-boundary cases pushes the model to resolve subtle inter-class differences, which is critical in fine-grained retrieval. Large batches enlarge the candidate pool, raising the chance of sampling truly hard pairs and yielding faster convergence and more discriminative embeddings.
Additionally, we use a Straight-Through Estimator (STE) with a low-temperature softmax ($\tau=0.05$) for the non-differentiable $\arg\max$ in the ROI buddy-patch selection during backpropagation. In the forward pass, the selected buddy patch remains the hard maximum; in the backward pass, the low-temperature softmax provides an approximate gradient through the selection scores. We use this setting in all reported experiments. The temperature affects only the backward pass, and at inference buddy selection uses a strict hard $\arg\max$.

\paragraph{Data Augmentation}
We apply multi-scale padding augmentation only to INSTRE to reduce the domain shift between cropped queries and full gallery images.
Each cropped query is centered on a fixed-size black canvas with a binary mask $\mathbf{M}\in\{0,1\}^{h\times w}$ marking real content; during buddy selection and ROI pooling (Eq.~\ref{eq:buddy_avg}) only positions with $\mathbf{M}_{u,v}=1$ contribute, preventing border regions from biasing similarities or diluting the ROI. Presenting the object at multiple scales better matches gallery scales, and at evaluation we keep the maximum similarity across scales.

\subsection{Results and Ablations}
\label{subsec:results_ablations}

Table~\ref{tab:resultscomparison} compares our approach (with $R=4$ register tokens and one ROI token per cue token) to several other methods. Since these methods use different backbones, pretraining schemes, and training objectives, this table should be read as a comparison to the reported literature rather than as an isolated attribution of our architectural choices. We isolate the effect of registers, ROI tokens, and multi-vector scoring in the DINOv2-reg ablations below.  
Since each configuration requires retraining, the ablations in
Tables~\ref{tab:ablation_register} and~\ref{tab:roi_size_ablation}
are restricted to CUB and Cars-196, the two smaller 
datasets.
The main insights from our ablations are: \textit{(i)} the compact multi-vector representation improves over the DINOv2-reg \texttt{[CLS]} baseline in our evaluated settings; and \textit{(ii)} ROI window size affects performance, with $3\times3$ providing a good default in our experiments.

\paragraph{Results on small objects (INSTRE dataset)}
Table~\ref{tab:instre_results} reports mAP on the challenging INSTRE dataset, whose small targets, cluttered scenes, and strong scale, viewpoint, and illumination variation make instance-level discrimination hard. Among the baselines, SSLeb (Edge Boxes) and SSLss (Selective Search) differ in the region-generation strategy and the +Bd variants add a diffusion-based self-boosting refinement stage. Our method obtains higher mAP than the listed baselines.

\begin{table}[h]
\centering
\caption{Comparison of retrieval performance (mAP) on the \mbox{INSTRE} dataset against state-of-the-art methods.}
\label{tab:instre_results}
\resizebox{.46\textwidth}{!} {
\begin{tabular}{cccccccccc}
\toprule
R-MAC~\cite{tolias2015particular} & GeM~\cite{radenovic2019} & GeM+AP~\cite{revaud2019} & SSLeb~\cite{zhang2022instanceIRL} \\
\midrule
43.0\% & 69.0\% & 33.5\% & 92.2\% \\
\bottomrule
\toprule
SSLss~\cite{zhang2022instanceIRL} & SSLeb+Bd~\cite{zhang2022instanceIRL} & SSLss+Bd~\cite{zhang2022instanceIRL} & \textbf{Ours} \\
\midrule
92.0\% & 92.5\% & 92.3\% & \textbf{93.5\%} \\
\bottomrule
\end{tabular}
}
\end{table}

\paragraph{Register vs.\ non-registers}
In Table \ref{tab:ablation_register}, we compare using DINOv2-reg with only the \texttt{[CLS]} token, with \texttt{[CLS]} plus register tokens, and with the full \texttt{[CLS]}+register+ROI representation. Register tokens alone provide a modest gain, especially on Cars-196, while the buddy-pooled ROI tokens provide the larger improvement on CUB and a further gain on Cars-196. These results suggest that registers are useful complementary cues, but that the main benefit of our compact multi-vector representation comes from combining them with localized ROI tokens.

\begin{table}[!ht]
\centering
\footnotesize
\caption{Effect of using register and ROI tokens (bottom 2 experiments using multi-vector training).}
\label{tab:ablation_register}
\resizebox{.48\textwidth}{!} {
\begin{tabular}{lcc}
\toprule
Method & CUB (R@1) & Cars-196 (R@1) \\
\midrule
DINOv2-reg (\texttt{[CLS]} only) & 86.6 & 89.8\\
DINOv2-reg (\texttt{[CLS]}+Registers) & 86.7 & 90.3 \\
DINOv2-reg (\texttt{[CLS]}+Registers+ROI)       & \textbf{87.4} & \textbf{90.5} \\
\bottomrule
\end{tabular}
}
\end{table}

\paragraph{Single-vector vs.\ Multi-vector training}
Next, we evaluate the importance of multi-vector training instead of only single-vector training. Both models are trained identically until the scoring; we compare the single-vector used in the triplet loss, to our approach, where we aggregate the scores from the \texttt{[CLS]}, register, and ROI tokens via a ColBERT-style loss. As shown in Table~\ref{tab:ablation_register}, the single-vector setting underperforms the multi-vector method. This demonstrates how we utilize the ROI tokens during training is critical for local alignment, especially in FGIR tasks.
Furthermore, because these tokens already exist as a part of the forward pass and computing ROI tokens requires little additional computation, our method offers an easy drop-in replacement in existing retrieval pipelines. 

\paragraph{Ablation on region size for ROI tokens}
In Table \ref{tab:roi_size_ablation}, we evaluate the effect of the ROI window size. The results suggest that pooling a local neighborhood around the buddy patch is preferable to using only the single selected patch, while larger windows provide limited additional benefit and may depend on the dataset. We therefore use a $3\times3$ ROI window as the default setting in the remaining experiments.

\begin{table}[!h]
\centering
\caption{
  Ablation on the region size for ROI tokens.
  We report Recall@1 on CUB and Cars-196 with
  single patch vs. $N \times N$ mean pooling for N=3, 5, 7, 9. Our default setting is N=3.
}
\label{tab:roi_size_ablation}
\resizebox{.36\textwidth}{!} {
\begin{tabular}{lcc}
\toprule
\textbf{ROI Size} & \textbf{CUB (R@1)} & \textbf{Cars-196 (R@1)} \\
\midrule
Single Patch & 87.8 & 89.4 \\$3\times3$ & 88.1 & 90.1 \\  $5\times5$ & 88.1 & 90.5 \\  $7\times7$ & 88.2 & 90.0 \\  $9\times9$ & 87.8 & 90.1 \\  \bottomrule
\end{tabular}
}
\end{table}

\paragraph{Memory Consumption vs.\ Performance} In Table~\ref{tab:memory_comparison}, we compare Recall@1 measured on CUB-200 with analytically projected raw float32 descriptor storage for a hypothetical database of 1M images. The projection illustrates storage scaling only and should not be interpreted as an evaluation of retrieval performance at the million-image scale.
The \textit{All Patch+Reg+\texttt{[CLS]}} row corresponds to the full-token
reference in which all $P$ patch tokens, the four register tokens, and the
\texttt{[CLS]} token (201 tokens per image) are matched directly, without
buddy-based ROI construction.
Our method uses more storage than a single-vector representation, but far less than storing all patch tokens, while retaining most of the observed benefit of compact multi-vector matching for fine-grained image retrieval. This comparison does not include indexing overhead, query latency, batching effects, or ANN-specific implementation costs; integrating and benchmarking such indexing schemes remains future work.

\begin{table}[!ht]
\footnotesize
\centering
\caption{Theoretical index size comparisons for 1M images using 384-dimensional float32 embeddings (4 bytes/dim). Recall@1 is measured on CUB-200 and does not represent million-scale retrieval performance. ``T. Dim'' and ``Memory'' denote dimensions per image and raw storage without overhead, respectively.}
\label{tab:memory_comparison}
\resizebox{.48\textwidth}{!} {
\begin{tabular}{lcccc}
\toprule
\textbf{Method} &
\textbf{\# Tokens} &
\textbf{T. Dim} &
\textbf{Memory} &
\textbf{Recall@1}\\
\midrule
Single-Vector &
1 &
384 &
1.5\,GB &
86.6 \\

\texttt{[CLS]}+Registers &
5 &
1{,}920 &
7.7\,GB &
86.7 \\

Ours: \texttt{[CLS]}+Reg+ROI &
10 &
3,840 &
15\,GB &
87.4 \\ 
All Patch+Reg+\texttt{[CLS]} &
201 &
77{,}184 &
309\,GB &
87.3 \\
\bottomrule
\end{tabular}
}
\end{table}

\paragraph{Comparison with baselines based on local features}

Table~\ref{tab:local_res} compares our results with those reported in the respective papers for recent local feature-based re-ranking methods. These approaches are typically designed for instance-level image retrieval, since they exploit local correspondences to better discriminate between specific object instances rather than broad categories. Unlike RRT~\cite{Tan2021} and LOCORE~\cite{xiao2025locore}, whose learned local-feature re-ranking cost depends on the number of re-ranked candidates and retained local descriptors, our approach constructs a fixed set of 10 tokens within a single backbone forward pass, without additional parameters or forward passes (Section~\ref{subsec:colbert_training}). Our method obtains the highest R@k in five of the six reported comparisons, suggesting that the selected register and ROI tokens provide useful local information for retrieval.

\begin{table}[!h]
\centering
\caption{Comparison of our approach with local feature-based methods on metric learning benchmarks in terms of Recall@k.}
\label{tab:local_res}
\small
\resizebox{0.48\textwidth}{!}{\begin{tabular}{lcccccc}
\toprule
& \multicolumn{3}{c}{CUB-200} & \multicolumn{3}{c}{In-Shop} \\
\cmidrule(lr){2-4} \cmidrule(lr){5-7}
& R@1 & R@2 & R@4 & R@1 & R@10 & R@20  \\
\midrule
RRT~\cite{Tan2021}          & 68.7 & 85.0 & 95.6 & 88.3 & 97.9 & 98.6 \\
LOCORE-tiny~\cite{xiao2025locore}        & 71.4 & 86.8 & 96.4 & 89.1 & 97.9 & 98.2 \\
LOCORE-small~\cite{xiao2025locore}       & 74.6 & 89.1 & 97.3 & 89.4 & 97.7 & 97.7 \\
LOCORE-base~\cite{xiao2025locore}        & 78.3 & 91.9 & \textbf{98.2} & 87.9 & 97.9 & 98.7 \\
\hline
\textbf{Ours} & \textbf{87.4} & \textbf{92.7} & 95.2 & \textbf{92.9} & \textbf{98.4} & \textbf{98.9} \\
\bottomrule
\end{tabular}}
\end{table}

\section{Conclusion and Future Work}
\label{sec:conclusion}

We show that DINOv2 register tokens provide complementary information for image retrieval. Our framework combines the \texttt{[CLS]} token, four register tokens, and ROI tokens obtained by buddy pooling in a compact multi-vector representation of 10 tokens per image. In the evaluated settings, this representation improves over the corresponding DINOv2-reg single-vector baseline, with the largest gains arising from the ROI tokens. It requires no external modules and uses only the backbone's internal representations. Our ablations further analyze the ROI size and token composition.

For future work, we plan to compare our ROI selection guided by cue tokens with attention, random selection, and attention across images.
We also plan to evaluate configurations without training and with fewer tokens (e.g., \texttt{[CLS]} plus its most frequently matched register), and to extend the mechanism to CLIP, DINOv3 (unavailable when this study started), and registers added at inference.

\section*{Acknowledgment}

We thank the São Paulo Research Foundation – FAPESP (grant \#2025/10602-5), the University of São Paulo (PRPI Ordinance No. 1032, New Faculty Support Program), and the U.S. National Science Foundation (NSF) award IIS-2331768.

\end{document}